\documentclass[11pt]{article}

\usepackage[margin=2.9cm]{geometry}
\RequirePackage[colorlinks,citecolor=blue,urlcolor=blue]{hyperref}

\usepackage[citestyle=alphabetic,
	backend=bibtex,
	doi=false,
	isbn=false,
	url=false,
	firstinits=true,
	]{biblatex}

\renewbibmacro{in:}{}			
\DeclareFieldFormat[article,inbook,incollection,inproceedings,patent,thesis,unpublished]{citetitle}{#1}
\DeclareFieldFormat[article,inbook,incollection,inproceedings,patent,thesis,unpublished]{title}{#1} 

\usepackage{amssymb}
\usepackage{amsmath}
\usepackage{amsthm}
\usepackage{graphicx}
\usepackage{bm}
\usepackage[inline]{enumitem}
\usepackage{mathrsfs}

\usepackage[font=small]{caption}

\RequirePackage{algorithm}
\RequirePackage{algpseudocode}

\RequirePackage{xpatch}
\makeatletter
\xpatchcmd{\algorithmic}{\itemsep\z@}{\itemsep=.5ex plus2pt}{}{}
\makeatother
\algnewcommand{\IIf}[1]{\State\algorithmicif\ #1\ \algorithmicthen}

\bibliography{ref-LL-AA}

\usepackage[matrixnorms,opnorm=op]{arash_macros}

\title{Exact slice sampler for Hierarchical Dirichlet Processes} 
\author{Arash A. Amini, Marina  Paez, Lizhen Lin and Zahra S. Razaee}
\newcommand\pib{\bm{\pi}}
\newcommand\betab{\bm{\beta}}
\DeclareMathOperator\gem{GEM}
\DeclareMathOperator\Beta{Beta}
\DeclareMathOperator\DP{DP}
\DeclareMathOperator\unif{Unif}
\newcommand\gamb{\bm{\gamma}}
\newcommand\kb{\bm{k}}
\newcommand\tb{\bm{t}}
\newcommand\xb{\bm{x}}
\newcommand{\zb}{\bm{z}}
\newcommand\n{n}
\newcommand\ub{\bm{u}}
\newcommand\vb{\bm{v}}
\newcommand{\Fc}{\mathscr F}
\newcommand\fb{\bm{f}}
\newcommand\yb{\bm{y}}

\newcommand{\alphab}{\bm\alpha}

\begin{document}

\maketitle

\begin{abstract}
	
	We propose an exact slice sampler for Hierarchical Dirichlet process (HDP) and its associated mixture models~(\citeauthor*{hdp}, \citeyear{hdp}).  Although there are existing MCMC algorithms for sampling from the HDP, 
	a slice sampler has been missing from the literature. Slice sampling is well-known for its desirable  properties including its fast mixing  and its natural potential for parallelization. On the other hand, the hierarchical nature of HDPs poses challenges to adopting a full-fledged slice sampler that automatically truncates all the infinite measures involved without ad-hoc modifications.  In this work, we adopt the powerful idea of Bayesian variable augmentation to address this challenge. By introducing new latent variables, we obtain a full factorization of the joint distribution that is suitable for slice sampling. Our algorithm has several appealing features such as (1) fast mixing; (2) remaining exact while allowing natural truncation of the underlying infinite-dimensional measures, as in~(\citeauthor*{Kalli11}, \citeyear{Kalli11}), resulting in updates of only a finite number of necessary atoms and weights in each iteration; and (3) being naturally suited to parallel implementations.  The underlying principle for  joint factorization of the full likelihood is simple and can be applied to many other settings, such as designing sampling algorithms for general dependent Dirichlet process (DDP) models. 
	
\end{abstract}

\section{Introduction}

Hierarchical Dirichlet process (HDP)~\cite{hdp} is one of the popular Bayesian nonparametric models for modeling the hierarchy of groups of data. It has been widely applied in various learning tasks in statistics and machine learning, such as topic modeling~\cite{hdp,Newman:distri-topic} and information retrieval~\cite{hdp}. It is often used as the mixing measure in a mixture model  for modeling the cluster structure for groups of data~\cite{sohn2009}.  Other applications include using HDP  for modeling the transition probabilities between hidden states in a hidden Markov model~\cite{fox2011}.
Being able to sample efficiently from a HDP is crucial for making inference in HDP-related models, and to do so,  both approximating algorithms and sampling-based algorithms have been proposed in the literature.  Approximating algorithms for HDP are mostly  based on variational approaches~\cite{kurihara:variation,Adistri-topic,Newman:distri-topic}. These algorithms can often scale to large datasets but suffer from some obvious drawbacks: the variational posterior tends to underestimate the variability of the true posterior.

In this work, we will focus on the sampling-based methods. One of the prominent methods is the Chinese Restaurant Franchise (CRF)-based Gibbs sampler \cite{hdp}. This algorithm, however, is known to mix slowly,  encouraging the search for more efficient sampling  approaches. Attempting that, \cite{fox2011} adopted a truncated approximation to the full posterior distribution for sampling HDP in a hidden Markov model, and \cite{wang-blei} proposed a split/merge MCMC algorithm, which, however, according to~\cite{chang2014parallel} only shows a marginal improvement over the sampler of~\cite{hdp}. More recently, \cite{chang2014parallel} proposed an algorithm that also considers split/merge steps but has the advantage of allowing parallel sampling, and was reported to exhibit significantly improved convergence  compared to~\cite{wang-blei}. In~\cite{online-hdp}, online algorithms based on mini-batch ideas are proposed.

An efficient approach to sample from DP mixtures is through slice sampling. This approach has the advantage of allowing  natural truncation of  the infinite-dimensional Dirichlet measures, as shown in the seminal work of~\cite{Kalli11}. Slice samplers are also known to have great  mixing properties. To the best of our knowledge, however,   no slice sampler  has been proposed for HDPs. We fill this gap with the development of an \emph{exact slice sampler} for HDP, extending the ideas of the slice sampling for DPs to the hierarchical setup. 

One difficulty in performing slice sampling for HDPs arises from the hierarchical nature of the HDP model. We  refer to Section~\ref{sec:label:only} for more details on the difficulty facing the hierarchical stick-breaking representation and the marginalization approach. To circumvent this, we adopt the powerful idea of Bayesian variable augmentation  which allows a full factorization of the joint distribution. The slice sampler that we derive for HDPs retains the same advantage of being exact while truncating the underlying infinite-dimensional measures ($\betab$ and $\gamb_j$ in our notation). In other words, the sampler only updates the necessary atoms and weights in each iteration, whose number is guaranteed to be finite. 

The slice sampler proposed here enjoys several key features: it is very simple and its updates are fairly intuitive. It is also naturally parallel. 
 Moreover, as a by-product, we derive a  complete factorization of the  joint density of HDP that could be of independent interest. The expression for the joint density allows one to easily verify the validity of the slice sampler updates; i.e., not much knowledge of Dirichlet processes and their intricacies is required, and the derivations are accessible to most practioners with a basic understanding of Bayesian statistics. 

Our motivation for this new algorithm came from a network point of view, and the need to propose inference models for multiplex networks that can take into account potential dependency across different layers, particularly when the aim is community detection. In this work (that is currently in progress), we specify HDP as a natural random partition prior for the partitions across different layers in the multiplex network. Despite our original motivation, it is important to point out that our algorithm can be easily generalized to other models if one replaces the mixture part with a general likelihood, therefore immediately getting a nicer sampling scheme.

\section{Representation of HDPs}
We start by deriving a representation of the HDPs which is suitable for slice sampling. We assume familiarity with the setup of a HDP as in~\cite{hdp} and the associated metaphor of a Chinese restaurant franchise (CRF). We first focus on the label generation part of the HDP, and then discuss how one can add the bottom mixture layer.

\subsection{Label-only HDP}\label{sec:label:only}

Consider the HDP as defined in Equ.~(19) of~\cite{hdp}. Using mostly the same notation %
and the CRF metaphor, we write
\begin{align*}
\betab \mid \gamma_0 &\sim \gem(\gamma_0) \\
\pib_j \mid \alpha_o, \betab &\sim \DP(\alpha_0,\betab), \\ 
z_{ji} \mid \pib_j &\sim \pib_j
\end{align*}
where $i=1,\dots,n_j$ is the customer index and $j =1,\dots,J$ is the restaurant index. We are ignoring the downstream mixture model for the moment, since this is the main part of the sampling problem. Thus, we assume that we observe the labels $\{z_{ji}\}$ where $z_{ji}$ is the dish of customer $i$ in restaurant $j$, and we want to estimate $\betab$ and $\{\pib_j\}$.

The stick-breaking representation of $\pib_j$---see Equ.~(21) in~\cite{hdp}---is not suitable for slice (or Gibbs) sampling due to the complicated dependence on $\betab$. Another idea is to marginalize $\pib_j$, but that would lead to distributions with ratios of Gamma functions as densities which are not easy to sample from. Instead, we just use the fact that $\pib_j$ is itself a Dirichlet measure. That is, we can write the model as
\begin{align*}
\betab \mid \gamma_0 &\sim \gem(\gamma_0) \\
\gamb_j \mid \alpha_0 &\sim \gem(\alpha_0), \quad \gamb_j = (\gamma_{jt}) \\
k_{jt} \mid \betab &\sim \betab, \quad t \in \nats \\
\pib_j &= \sum_{t=1}^\infty \gamma_{jt} \delta_{k_{jt}}, \quad 	z_{ji} \mid \pib_j \sim \pib_j.
\end{align*}
Note that $\gamb_j$ and $(k_{jt})$ are independently drawn. Here, index ``$t$'' is interpreted as indexing the tables; $k_{jt}$ is the dish (type) of table $t$ in restaurant $j$, while $\gamma_{jt}$ represents the fraction of the customers in restaurant $j$ that would sit at table $t$ (eventually).

This representation is still not suitable for sampling. Instead of sampling directly from $\pib_j$, we sample from the weights $\gamb_{j}$ first, i.e., we pick the table of customer $i$  and then assign them the dish of the table. This gives us access to the last missing piece which is $t_{ji}$, the table of customer $i$ in restaurant~$j$. We can write the model equivalently as
\begin{align}\label{eq:label:only:HDP:v3}
\begin{split}
\betab \mid \gamma_0 &\sim \gem(\gamma_0), \quad \betab = (\beta_k) \\
\gamb_j \mid \alpha_0 &\sim \gem(\alpha_0), \quad \gamb_j = (\gamma_{jt}) \\
k_{jt} \mid \betab &\sim \betab, \quad t \in \nats, \\ 
t_{ji} \mid \gamb_j &\sim \gamb_j, \quad i=1,\dots,n_j \\
z_{ji} \mid \tb_j, \kb_j &= k_{j,t_{ji}} 
\end{split}
\end{align}
where $\kb_j = (k_{jt}, t \in \nats)$ is the collection of all the dishes at restaurant $j$. Note that we sample $t_{ji}$ (which table to sit customer $i$ in restaurant $j$) from the eventual distributions of customers among tables in restaurant $j$, i.e., $\gamb_j$. Equation $z_{ji} = k_{j,t_{ji}}$
means that the dish of customer $i$ in restaurant $j$, i.e. $z_{ji}$, is completely determined by looking at which table they are sitting at, $t_{ji}$, and what dish is presented at that table $k_{j,t_{ji}}$.

 Since given everything else, $z_{ji}$ is deterministic, we only need to worry about sampling $\betab$, $\gamb_j$, $\kb_j$ and $\tb_j = (t_{ji})$.
Let us define $F :[0,1]^\nats \to [0,1]^\nats$ by
\begin{align}\label{eq:F:def}
[F(\xb)]_1 :=  x_1, \quad [F(\xb)]_j := x_j \prod_{\ell=1}^{j-1} (1-x_\ell)
\end{align}
where $\xb = (x_j,j \in \nats)$. Both $\betab$ and $\gamb_j$ have stick-breaking representations~\cite{sethuraman94,ishwaran}:
\begin{align*}
\gamma'_{jt} &\sim \Beta(1,\alpha_0), \quad \beta'_k \sim \Beta(1,\gamma_0), \\
\gamb_j &= F(\gamb'_j), \qquad \qquad  \betab = F(\betab'), 
\end{align*}
where $\gamb'_j = (\gamma'_{jt})$ and $\betab' = (\beta'_k)$. Let us write $x \mapsto b_{\alpha_0}(x)$ for the density of $\Beta(1,\alpha_0)$, that is, $b_{\alpha_0}(x) \propto (1-x)^{\alpha_0-1}$.

Note that $\pr(t_{ji} = t) = \gamma_{jt},\; t \in \nats$.
Thus, we can write down the joint density as
\begin{align}\label{eq:joint:v1}
\begin{split}
p(\tb,\kb,\gamb',\betab') &= \prod_{j=1}^J \Big[ p(\tb_j | \gamb_j) \,  p(\gamb'_j) \,p(\kb_j | \betab) \, \Big] p(\betab') \\
&= \prod_{j=1}^J \left( \prod_{i=1}^{n_j} \gamma_{j,t_{ji}} \prod_{t=1}^\infty b_{\alpha_0}(\gamma'_{jt})\prod_{t=1}^\infty \beta_{k_{jt}} \right) \prod_{k=1}^\infty b_{\gamma_0}(\beta'_k).
\end{split}
\end{align}
Interestingly, this decomposition works for any other stick-breaking distributions on $\betab$ and $\gamb_j$. Using $\gamb_j = F(\gamb'_j)$ and $\betab = F(\betab')$, a more explicit formula is
\begin{align}\label{eq:joint:v2}
p(\tb,\kb,\gamb',\betab') = \prod_{j=1}^J \left( \prod_{i=1}^{n_j} [F(\gamb'_{j})]_{t_{ji}} \prod_{t=1}^\infty b_{\alpha_0}(\gamma'_{jt})\prod_{t=1}^\infty [F(\betab')]_{k_{jt}} \right) \prod_{k=1}^\infty b_{\gamma_0}(\beta'_k)
\end{align}
which gives the complete joint density of HDP in~\eqref{eq:label:only:HDP:v3}.

\subsection{Mixture part}
Finally, we can add in the mixture component as
\begin{align}\label{eq:mix:comp}
\begin{split}
f_k \mid \Fc &\sim \Fc, \quad \fb = (f_k) \\
y_{ji} \mid z_{ji}, \fb &\sim f_{z_{ji}}, 
\end{split}
\end{align}
where $\fb$ is an infinite collection of possible mixture components, where each coordinate $f_k$ is a density drawn from a distribution $\Fc$ on densities. We can assume $f_k = K(\cdot,\phi_k)$ for some kernel $K$, where $\phi_k \mid H \sim H$, to get back the more common parametric mixture model; the more general setup however is easier to work with conceptually. 

\begin{figure}[t]
	\centering
	\includegraphics[width=.52\textwidth]{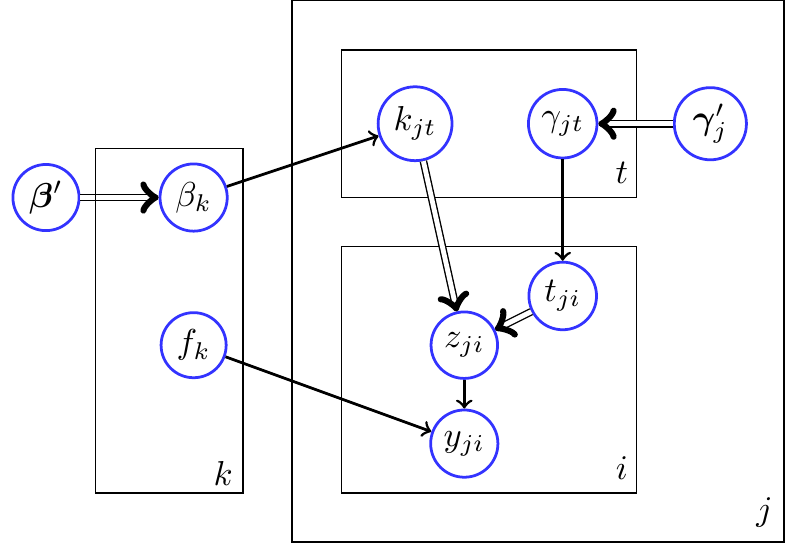}
	\caption{Schematic diagram of the HDP mixture with latent variables introduced for sampling. Double arrows show deterministic relations.}
	\label{fig:hdp}
\end{figure}

\medskip
Since $p(\yb \mid \tb,\kb,\fb) = \prod_{j=1}^J \prod_{i=1}^{n_j} f_{z_{ji}}(y_{ji})$, the overall joint density is
\begin{align}\label{eq:full:joint:v1}
\begin{split}
p(\yb,\fb,\tb,\kb,\gamb',\betab') &= p(\yb \mid \tb,\kb ,\fb)  \,p(\tb,\kb,\gamb',\betab') \,p(\fb) \\
&= \prod_{j=1}^J \left( \prod_{i=1}^{n_j} \big[ f_{  k_{j, t_{ji}} }(y_{ji}) \,\gamma_{j,t_{ji}} \big]  \prod_{t=1}^\infty b_{\alpha_0}(\gamma'_{jt})\prod_{t=1}^\infty \beta_{k_{jt}} \right) \prod_{k=1}^\infty \big[ b_{\gamma_0}(\beta'_k) \,\Fc(f_k)\big].
\end{split}
\end{align}
We note that this joint density is completely factorized over all its variables. The diagram of the model is shown in Figure~\ref{fig:hdp}.

\section{Sampling}

For the most part, when sampling, we can ignore the mixture part. That is, for the most part it is enough to sample from~\eqref{eq:joint:v2}. Only in sampling $\tb$ and $\kb$ the mixture part comes in. The factorized form of the density in~\eqref{eq:full:joint:v1} allows us to easily derive the Gibbs updates.

First, we state a key lemma. %
Let $x \mapsto b(x; \alpha,\beta)$ be the density of $\Beta(\alpha,\beta)$. %
The derivations in this paper can be extended to any stick-breaking prior for which a conjugacy relation similar to the one described in the lemma holds:
\begin{lem}\label{lem:basic:dp:slice}
	Assume that the joint density of $\xb = (x_1,x_2,\dots) \in [0,1]^\nats$ is proportional to
	\begin{align*}
	\prod_{i=1}^n [F(\xb)]_{z_i} \prod_{j=1}^\infty b(x_j; \alpha, \beta),
	\end{align*}
	where $\zb = (z_1,\dots,z_n) \in \nats^n$ and $F : [0,1]^\nats \mapsto [0,1]^\nats$ is defined as in~\eqref{eq:F:def}. Then
	\begin{align*}
	x_j \mid \xb_{-j} \sim \Beta\big(\n_j(\zb)+\alpha, \n_{>j}(\zb)+\beta\big)
	\end{align*}
	where $	\n_j(\zb) = |\{i:\; z_i = j \}|$ and $\n_{>j}(\zb) = |\{i:\; z_i > j \}|$.
\end{lem}
The proof is given in Appendix~\ref{app:useful:lemma}.

\subsection{Usual block Gibbs sampling}

\paragraph{Sampling $\gamb' \mid \tb,\kb, \betab'$.} This posterior factorizes over $\gamb'_j$, and the posterior of $\gamb'_j$ given the rest is proportional to $\prod_{i} [F(\gamb'_{j})]_{t_{ji}} \prod_{t} b_{\alpha_0}(\gamma'_{jt})$. Using Lemma~\ref{lem:basic:dp:slice},  we have
\begin{align}\label{eq:gamb:p:sampling}
\gamma'_{jt} \mid \gamb'_{-jt}, \tb,\kb, \betab' \sim \Beta\big(\n_t(\tb_j)+1, \n_{>t}(\tb_j)+\alpha_0\big).
\end{align}
Note that, for a fixed $j$, we are applying Lemma~\ref{lem:basic:dp:slice}  to the factorization indexed by $t$. Here, $n_t(\tb_j) = \{i:\; t_{ji} = t\}|$ and $n_{>t}(\tb_j) = |\{i:\; t_{ji} > t\}|$.

\paragraph{Sampling $\betab' \mid \tb,\kb, \gamb'$.} This posterior is proportional to $\prod_{j} \prod_t  [F(\betab')]_{k_{jt}}  \prod_{k} b_{\gamma_0}(\beta'_k)$. Applying Lemma~\ref{lem:basic:dp:slice} to the factorization over $k$, we obtain
\begin{align}\label{eq:betab:p:sampling}
\beta'_k \mid \betab'_{-k}, \gamb',  \tb,\kb \sim \Beta\big(\n_k(\kb)+1, \n_{>k}(\kb)+\gamma_0\big)
\end{align}
where $\n_k(\kb) = |\{(j,t) :\; k_{j,t} = k\} |$ and similarly for $n_{>k}(\kb)$.

\paragraph{Sampling $\tb \mid \kb, \gamb',\betab', \yb,\fb$.} This posterior factorizes over $j$ and $i$ and is given by 
\begin{align*}
t_{ji} \mid \gamb'_j, \kb_j, \yb_j,\fb &\;\sim\; \big( f_{  k_{jt}}(y_{ji}) \,\gamma_{jt} \big)_{t \in \nats}.%
\end{align*}

\paragraph{Sampling $\kb \mid \tb, \gamb',\betab', \yb,\fb$.} This posterior factorizes over $j$. Writing 
\begin{align*}
f_{  k_{j, t_{ji}} }(y_{ji}) = \prod_{t=1}^\infty \big[ f_{  k_{jt} }(y_{ji})  \big]^{1\{ t_{ji} = t\}},
\end{align*}
the posterior for $\kb_j$, given the rest, is proportional to 
\begin{align}\label{eq:kb:j:post}
\prod_{i=1}^{n_j} \big[ f_{  k_{j, t_{ji}} }(y_{ji})  \big]  \prod_{t=1}^\infty \beta_{k_{jt}} = 
\prod_{t=1}^\infty \Big( \beta_{k_{jt}}  \prod_{i=1}^{n_j}    \big[ f_{  k_{jt} }(y_{ji}) \big]^{1\{ t_{ji} = t\}} \Big)
\end{align}
which also factorizes over $t$. Thus, it is enough to sample (independently over $j$ and $t$),
\begin{align*}
k_{jt} \mid \betab', \tb_j, \yb_j,\fb &\;\sim\; \Big( \beta_k \prod_{i:\; t_{ij} = t} f_{  k }(y_{ji})  \Big)_{k \in \nats}.
\end{align*}

\paragraph{Sampling $\fb \mid \cdots$.} Recalling $z_{ji} = k_{j,t_{ji}}$, we sample independently over $k$,
\begin{align}\label{eq:f:sampling}
p(f_k \mid \cdots) \; \propto \; \Fc(f_k) \prod_{(i,j):\; z_{ji} = k} f_k(y_{ji}).
\end{align}

\subsection{Slice sampling}
We recall the basic idea of slice sampling, which itself is a form of variable augmentation: In order to sample from density $f(x)$, we introduce the nonnegative variable $u$, and look at the joint density $g(x,u) = 1\{u \le f(x)\}$ whose marginal over $x$ is $f(x)$. Then, we perform Gibbs sampling on the joint $g$. In the end, we only keep samples of $x$ and discard those of $u$. This idea has been successfully employed in~\cite{Kalli11} to sample from the classical DP mixture. We now extend the ideas in~\cite{Kalli11} to sample from HDP mixtures.

To carry the idea over to the HDPs, we augment the model by adding variables $\ub_j = (u_{ji})$ and $\vb_j = (v_{jt})$ and consider the joint density
\begin{align}\label{eq:aug:joint}
\begin{split}
&p(\yb,\fb,\tb,\kb,\gamb',\ub,\betab',\vb) = \\
&\qquad  \prod_{j=1}^J \left( \prod_{i=1}^{n_j}  f_{  k_{j, t_{ji}} }(y_{ji}) 1\{ u_{ji} \le \gamma_{j,t_{ji}}\} 
\prod_{t=1}^\infty b_{\alpha_0}(\gamma'_{jt})\prod_{t=1}^\infty 1\{v_{jt} \le \beta_{k_{jt}}\} \right) 
\prod_{k=1}^\infty \big[ b_{\gamma_0}(\beta'_k) \,\Fc(f_k)\big].
\end{split}
\end{align}
Note that by integrating out the variables $(u_{ji})$ and $(v_{jt})$, we get back original joint density~\eqref{eq:full:joint:v1}. The idea is that we sample $(\gamb',\ub)$ jointly given the rest of variables, and similarly for $(\betab',\vb)$.

\paragraph{Sampling $(\gamb',\ub) \mid \tb,\kb, \betab',\vb$.} First we sample $(\ub \mid \gamb', \tb,\kb, \betab',\vb)$ which factorizes and the coordinate posteriors are $p(u_{ji} \mid \gamb',\tb\dots) \propto  1\{ u_{ji} \le \gamma_{j,t_{ji}}\}$, that is
\begin{align*}
u_{ji} \mid \gamb',\tb, \kb, \betab',\vb \; \sim\;  \unif(0, \gamma_{j,t_{ji}}).
\end{align*}
Next we sample from $(\gamb' \mid \tb,\kb, \betab',\vb)$. This would be the same as~\eqref{eq:gamb:p:sampling}.

\paragraph{Sampling $(\betab',\vb) \mid \kb,\tb, \gamb',\ub$.} First we sample $(\vb \mid \betab', \kb,\tb, \gamb',\ub)$ which factorizes and the coordinate posteriors are $p(v_{jt} \mid \betab',\kb\dots) \propto  1\{v_{jt} \le \beta_{k_{jt}}\} $, that is
\begin{align*}
v_{jt} \mid \betab', \kb,\tb, \gamb',\ub \; \sim\;  \unif(0, \beta_{k_{jt}}).
\end{align*}
Next, we sample from $(\betab' \mid \kb,\tb, \gamb',\ub)$. This would be the same as~\eqref{eq:betab:p:sampling}.

\paragraph{Sampling $\tb \mid \cdots$.} This posterior also factorizes over $i$ and $j$. From~\eqref{eq:aug:joint}, we have
\begin{align}\label{eq:slice:t:conditional}
\pr(t_{ji} = t \mid \tb_{-ji},\kb,\gamb',\ub,\betab',\vb) \;\propto\; f_{  k_{jt} }(y_{ji}) 1\{ u_{ji} \le \gamma_{jt}\}.
\end{align}
Let $T_{ji} := T(\gamb_j;u_{ji}) := \sup\{t:\;  u_{ji} \le \gamma_{jt} \}$. According to the above, $t_{ji}$ given everything else will be distributed as
\begin{align*}
t_{ji} \mid \cdots \;\sim\; \big(\, f_{  k_{jt}}(y_{ji})  \,\big)_{t \,\in\, [T_{ji}]}. %
\end{align*}
In the CRF metaphor, $T_{ji}$ is the maximum table index ($t$) that customer $i$ in restaurant $j$ can hop to at current iteration. Note that different customers are allowed different ranges of tables for their wandering. The update for $\tb$ is an instance of how the slice sampler truncates an infinite measure. Due to the presence of the indicator in~\eqref{eq:slice:t:conditional}, only values of $t$ for which $\gamma_{jt} \ge u_{ji}$ lead to a nonzero probability. In other words, the support of distribution~\eqref{eq:slice:t:conditional} is contained in $[T_{ji}]$. %

\paragraph{Sampling $\kb \mid \cdots$.} This posterior also factorizes over $j$ and $t$. From~\eqref{eq:aug:joint}, the posterior for $\kb_j$ given the rest is proportional to the same expression~\eqref{eq:kb:j:post} but with $\beta_{k_{jt}}$ replaced with $1\{v_{jt} \le \beta_{k_{jt}}\}$. Thus, we have
\begin{align*}
\pr(k_{jt} = k \mid \cdots) \;\propto\; 1\{ v_{jt} \le \beta_{k}\}  \prod_{i:\; t_{ji} = t} f_{  k }(y_{ji}) .
\end{align*}
Let $K_{jt} := K(\betab;v_{jt}) := \sup\{k:\;  v_{jt} \le \beta_k \}$. According to the above, $k_{jt}$ given everything else will be  distributed as
\begin{align*}
k_{jt} \mid \cdots \; \sim \; \Big( \prod_{i:\; t_{ji} = t} f_{  k }(y_{ji})  \Big)_{k \,\in\, [K_{jt}]}. %
\end{align*}
In CRF metaphor, $K_{jt}$ is the maximum dish index ($k$) available for substitution at table $t$ in restaurant $j$ at current iteration. Again different tables in the same restaurant have potentially different options for dish exchange. The update of $\kb$ is another instance where the slice sampler is truncating the   infinite measures involved.

\paragraph{Sampling $\fb \mid \cdots$.} This will be the same as~\eqref{eq:f:sampling}.

\newcommand\xib{\bm{\xi}}
\newcommand\Tmax{T^{\text{cap}}_j}
\newcommand\Kmax{K^{\text{cap}}}
\begin{algorithm}[th!]%
	\caption{Slice sampler for HDP mixture}
	\label{alg:hdp:slice:sampler}
	\linespread{1.1}\selectfont
	\begin{algorithmic}[1]

		\State Initialize $\Tmax$ and $\Kmax$ to pre-specified values (say 10). %
		\State Initialize $\tb_j$ and $\kb_j$ to all-ones vectors.
		\State Initialize $(u_{ji})$ and $(v_{jt})$ to independent uniform variables.
		\While {not \texttt{CONVERGED}, nor maximum iterations reached}
		
		\For {$j=1,\dots,J$}
		\State Sample $\gamma'_{jt} \sim \Beta\big(\n_t(\tb_j)+1, \n_{>t}(\tb_j)+\alpha_0\big)$ for all $t \in [\Tmax]$. \label{step:samp:gamp}
		\State Let $[\gamb_j]_{1:\Tmax} \gets [F(\gamb'_j)]_{1:\Tmax}$.
		\State  Let $T_{ji} \gets \max\{t:\;  u_{ji} \le \gamma_{jt} \}, \forall i \in [n_j]$ and $T_j \gets \max_{i=1,\dots,n_j} T_{ji}$.
		\State \texttt{doubling}($T_j$,\;$\Tmax$)%
		
		\EndFor

		\State 	Sample $\beta'_k \sim \Beta\big(\n_k(\kb)+1, \n_{>k}(\kb)+\gamma_0\big)$ for all $k \in [\Kmax]$. \label{step:samp:betap}
		\State Let $[\betab]_{1:\Kmax} \gets [F(\betab')]_{1:\Kmax}$.
		\State Let $K_{jt} \gets \max\{k:\;  v_{jt} \le \beta_k \}, \forall t \in [\Tmax], \,j \in [J]$.
		\State Let $K_{j} \gets \max_t K_{jt}$, \; \text{and} \; $K \gets \max_j K_j$. \label{step:Kjt:update}

		\State \texttt{doubling}($K$,\;$\Kmax$)%
		
		\State Sample $f_k$ from density $p(f \mid \cdots) \; \propto \; \Fc(f) \prod_{(i,j):\; z_{ji} = k} f(y_{ji})$ for all $k \in [\Kmax]$. \label{step:sample:f}
		
		\For{$j=1,\dots,J$}

		\smallskip
		\State Sample $k_{jt} \sim \Big( \prod_{i:\; t_{ji} = t} f_{  k }(y_{ji})  \Big)_{k \,\in\, [K_{jt}]}$ for all $t \in [\Tmax]$. \quad Set $\kb_j \gets (k_{jt})$.
		\label{step:kb:update}

		\State Sample $v_{jt}  \; \sim\;  \unif(0, \beta_{k_{jt}})$ for all $t \in [\Tmax]$. \label{step:v:update}
		
		\smallskip
		\State Sample $t_{ji} \sim  \big(\, f_{  k_{jt}}(y_{ji})  \,\big)_{t \,\in\, [T_{ji}]}$ for all $i\in[n_j]$. \quad Set $\tb_j \gets (t_{ji})$.
		\State Sample $u_{ji}   \; \sim\;  \unif(0, \gamma_{j,t_{ji}})$ for all $i \in [n_j]$.
		
		\smallskip
		\State Set $z_{ji} \gets k_{j,t_{ji}}$.

		\EndFor

		\EndWhile
		
		\State \textbf{macro} \texttt{doubling}($K$,\,$\Kmax$)%
		\State \quad \textbf{if} $K < \Kmax$, \textbf{then}  continue \textbf{else} $\Kmax \gets 1.5\Kmax $ and go to the previous iteration.%
	\end{algorithmic}
\end{algorithm}

\subsubsection{How many atoms to keep?} Let us define
\begin{align*}
T_j := \max_{i=1,\dots,n_j} T_{ji}, \quad K_{j} := \max_{t=1,\dots,T_j} K_{jt}, \quad K := \max_{j} K_j
\end{align*}
so that $T_j$ determines the maximum table index ``$t$'' which we need to keep track of for restaurant $j$. Given $T_j$, one can compute $K_j$ which is the maximum number of dish index ``$k$'' we need to keep track of in restaurant $j$. Note that quantities $T_j$ and $K_j$ will be finite and random; they depend on $\gamb,\betab,\ub$ and $\vb$ and get updated in each iteration.

This completes the description of the slice sampler which is summarized in Algorithm~\ref{alg:hdp:slice:sampler}. The implementation, however, uses a few other ideas besides the update equations derived earlier. The difficulty is that the count parameters $T_{ji}$, $T_j$,  $K_{jt}$, $K_j$ and $K$ are interrelated among themselves and with other latent parameters of the model. Updating some of the parameters while keeping others fixed would create a chain of dependencies which is hard to track. 

The easiest way to assure that we always have sufficiently enough atoms, from all the infinite measures, is to put caps on their numbers, i.e., $\Tmax$ and $\Kmax$ in Algorithm~\ref{alg:hdp:slice:sampler}, and increase the cap whenever we hit it, and repeat the previous iteration (for which we need to keep record of the state of the chain one-step into the past). This is reflected in the \texttt{doubling} macro in Algorithm~\ref{alg:hdp:slice:sampler}. This procedure creates a little bit of redundancy, but after a few steps, the chain will remain within the caps, and one avoids resampling for all but a few early updates. The algorithm guarantees that we always have $T_j < \Tmax$ and $K < \Kmax$ and so the chain is sampling exactly. %

An advantage of the slice sampler is that all the updates in each step can be done in parallel over the underlying coordinates. Even updates at multiple steps involving disjoint sets of parameters can be performed in parallel.

\paragraph{Explicit mixture densities.} Using the more common notation $f_k(y) = K(y;\phi_k)$, with $\phi_k \mid H \sim H$, Step~\ref{step:sample:f} can be written as follows: Sample $\phi_k$ from
\begin{align}\label{eq:alt:phi:update}
p(\phi \mid \cdots) \; \propto \; H(\phi) \prod_{(i,j):\; z_{ji} = k} K(y_{ji}; \phi)\, \quad \text{ for all $k \in [\Kmax]$}
\end{align}
and set $f_k = K(\,\cdot\,;\phi_k)$.
Note that in Step~\ref{step:sample:f}, if the set $\{(i,j): z_{ji} = k\}$ is empty for some $k \in [K]$---which could happen since $z_{ji}$ has not yet been updated from the previous iteration  while $K$ has just been  updated---then the product evaluates to $1$, and we draw $f_k$ from the prior $\Fc(f)$ itself.

We also note that in updating $k_{jt}$ in Step~\ref{step:kb:update}, if the set $\{i:\; t_{ji} = t\}$ is empty, it means that customers are no longer sitting at table $t$. As before, we interpret  products over  empty sets as evaluating to $1$, hence the dish of the vacant table is updated uniformly at random. 

\begin{figure}[th!]
	\centering
	\includegraphics[width=.48\textwidth]{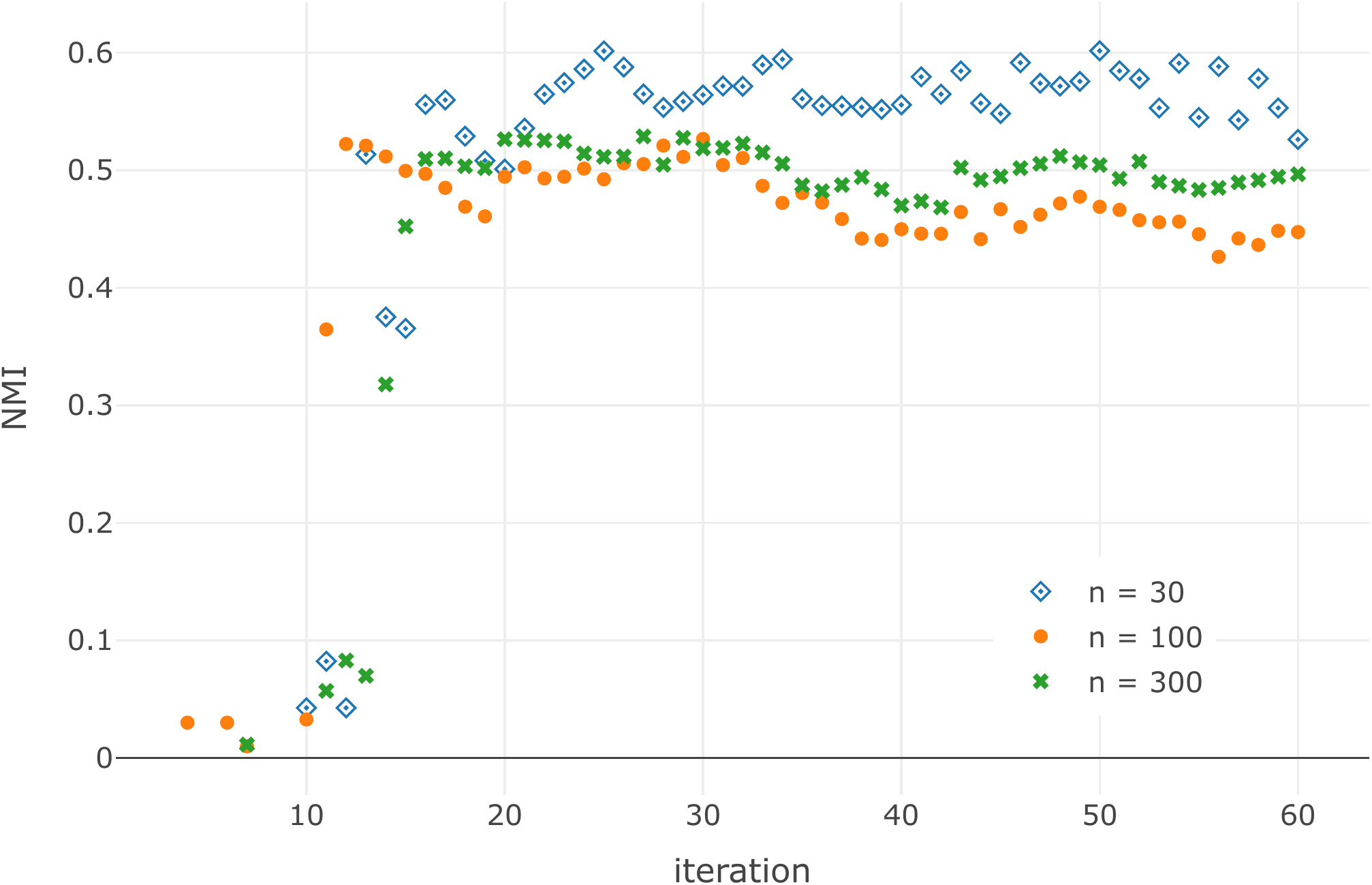}
	\includegraphics[width=.48\textwidth]{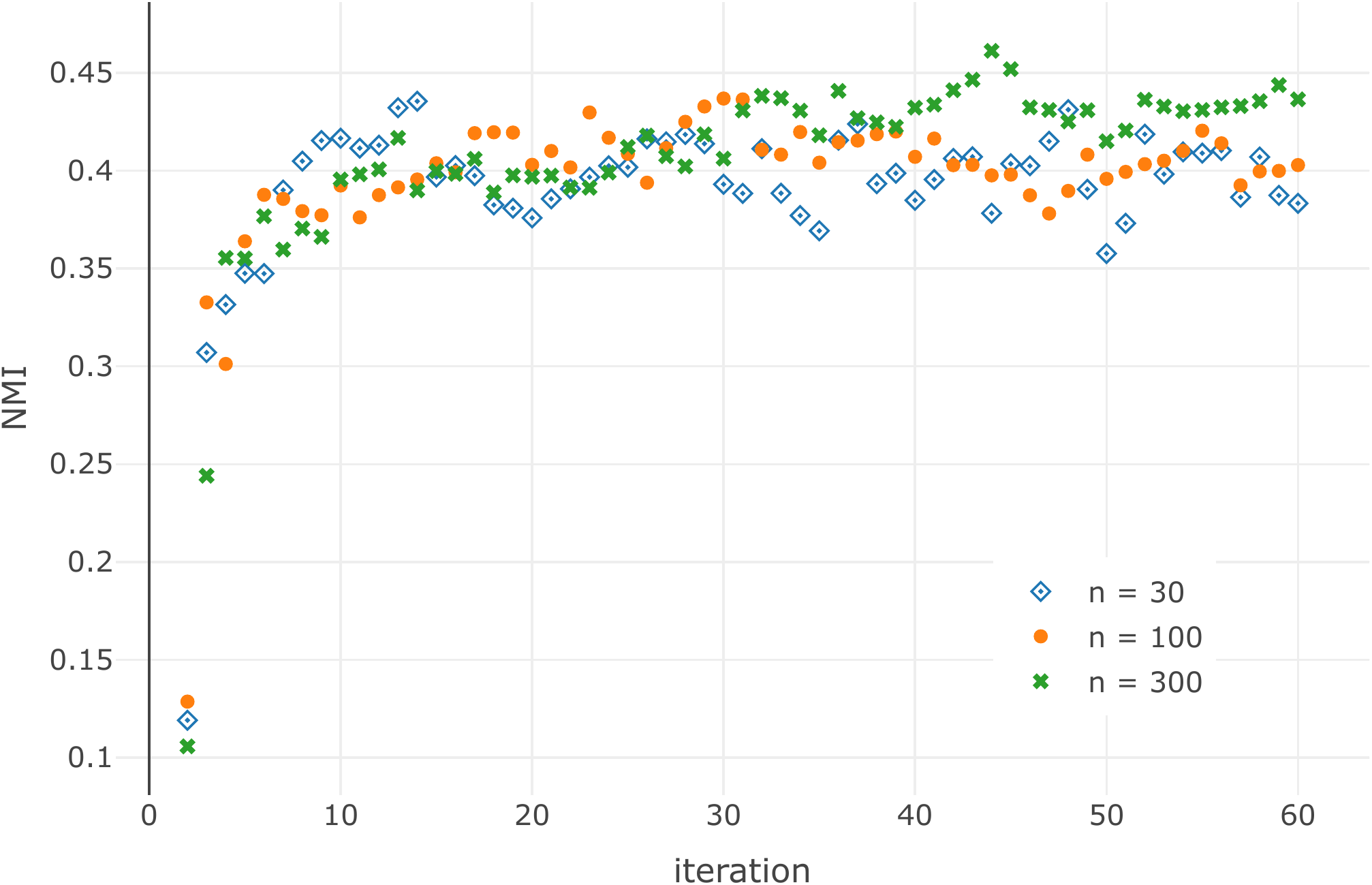}
	\includegraphics[width=.48\textwidth]{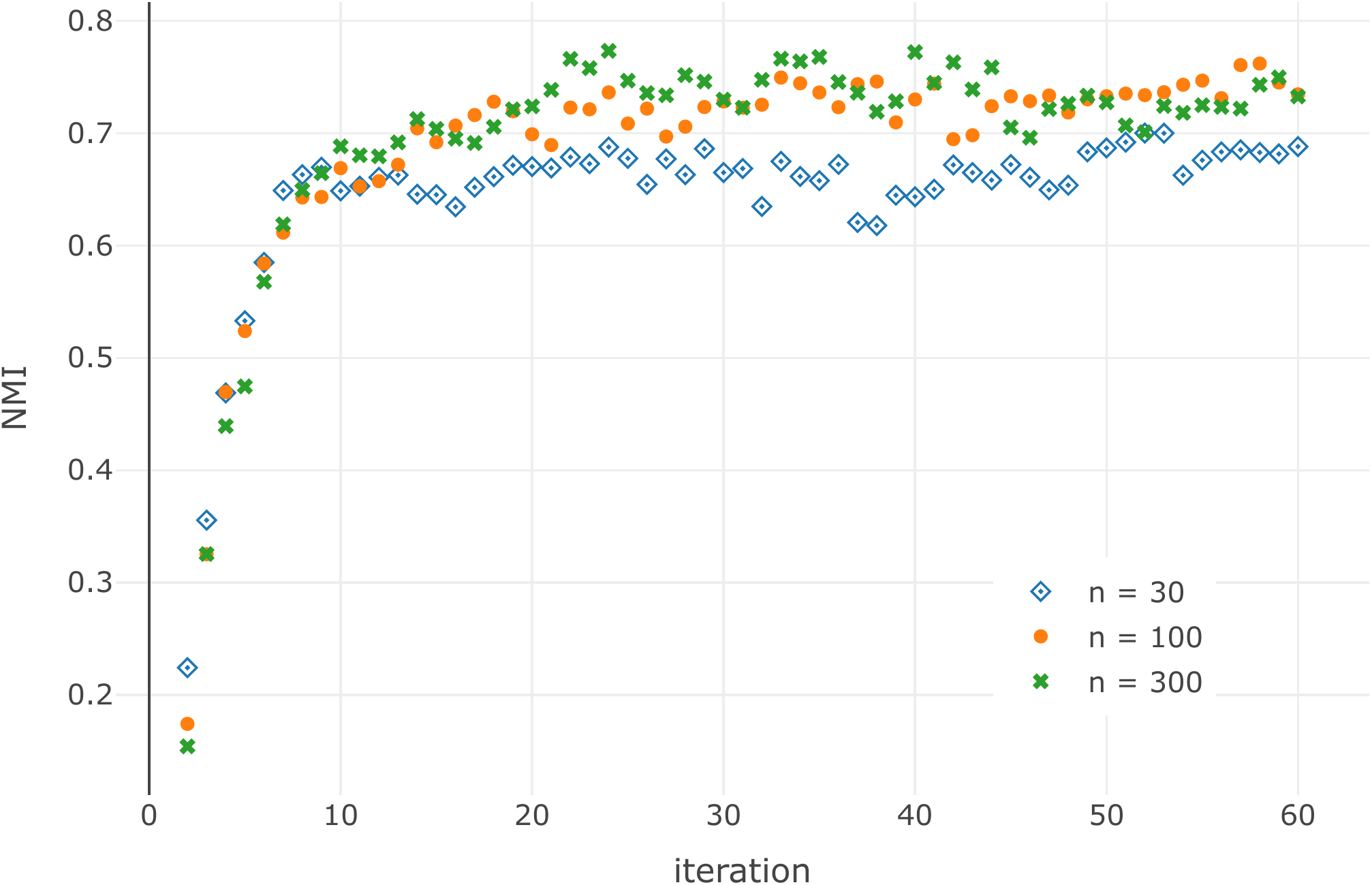}
	\includegraphics[width=.48\textwidth]{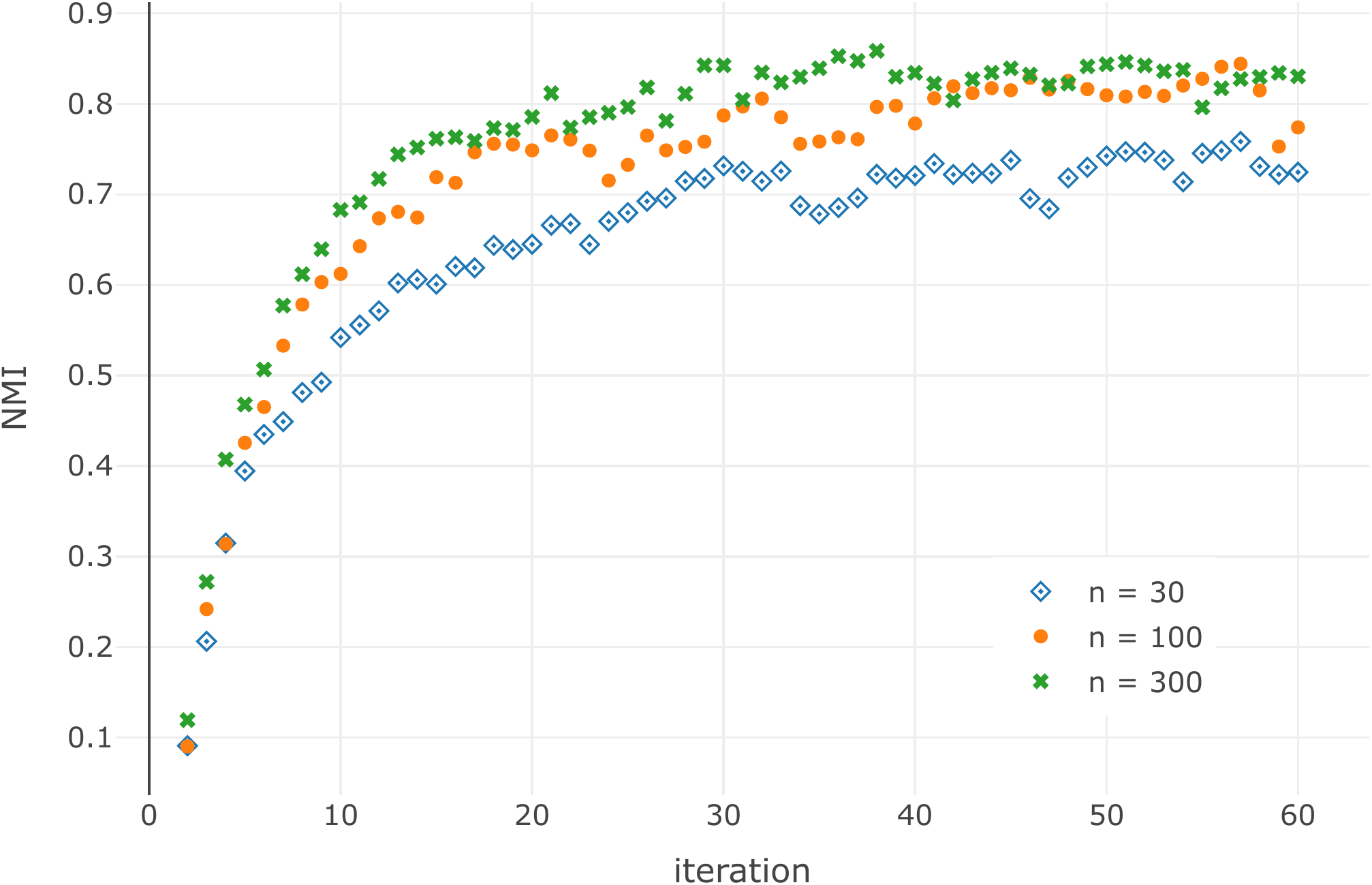}
	\caption{Typical Mixing behavior of the slice sampler for multinomial 
		HDP-mixtures for various sample sizes (Section~\ref{sec:experiments}). From top-left clockwise: $J=W=10,20,50,200$. Each plot shows the aggregate normalized mutual information (NMI) at each iteration. The NMI is computed for the labels from the posterior against the ``true'' labels. }
	\label{fig:mixing:1}
\end{figure}

\subsection{Examples}\label{sec:examples}
Let us consider a few examples of the mixture densities $f_k = K(\cdot;\phi_k)$. As a first example, consider a \emph{hierarchical Gaussian mixture}: We assume that 
\begin{align*}
	\phi_k &\mid H \sim H = N(0, I_d /\tau_y^2) \\
	y_{ji} &\mid z_{ji} \sim N(\phi_{z_{ji}},  I_d / \tau_y^2)
\end{align*}
where $\tau_\phi^2$ and $\tau_y^2$ are the precision parameters for the prior and the likelihood, respectively. For this model, it is not hard to see that the posterior update in Step~\ref{step:sample:f} is equivalent to drawing the atoms from a Gaussian distribution $\phi_k \mid \cdots  \;\sim\; N ( \mu_k,\tau_k^2 I_d)$ where
\begin{align}\label{eq:Gauss:update}
\mu_k = \frac{\tau_y^2 }{\tau_\phi^2 + n_k(\zb) \tau_y^2}  \sum_{(j,i) :\; z_{ji} = k} y_{ji}, \quad \text{and}\quad \tau_k^2 = \tau_\phi^2 + n_k(\zb) \tau_y^2.
\end{align}

As another example, consider a \emph{topic modeling} setup, where  $y_{ji}$ represents the $i$th word in document $j$. We assume $y_{ji} \in [W]$ where $[W] := \{1,2\dots,W\}$ is a vocabulary of $W$ words, identified with their index in a dictionary. Each atom $\phi \in [0,1]^W$ in this case represents a probability distribution over words in  vocabulary $[W]$. 
A natural prior on $\phi$ is $\text{Dir}((\alpha_w))$, i.e., 
\[
H(\phi) \propto \prod_{w=1}^W \phi_w^{\alpha_w-1},
\] 
and the likelihood is 
$
y_{ji} \mid z_{ji} \sim \text{Categorical}(\phi_{z_{ji}})
$ corresponding to the kernel 
\[K(y;\phi) = \phi_y= \prod_{w=1}^W \phi_w^{1 \{y=w\}}.\] 
The posterior update in Step~\ref{step:sample:f} (cf.~\ref{eq:alt:phi:update}) will be $\phi_k \mid \cdots \sim \text{Dir}(\alphab'_k)$ where %
$\alphab'_k$ has coordinates 
\[\alpha'_{kw} = \alpha_w + \sum_{j,i} 1\{y_{ji} = w, z_{ji} = k\}.\] 
We also note that  updating $\kb$---Step~\eqref{step:kb:update}-- simplifies to
\[
k_{jt} \sim \Big( \prod_w \phi_{kw}^{\nu'_{jtw}}\Big)_{k\, \in\, [K_{jt}]} \quad 
\text{where} \quad  \nu'_{jtw} = \sum_{i} 1\{y_{ji} = w, t_{ji} = t\}.
\]

\begin{figure}[th!]
	\includegraphics[width=.49\textwidth]{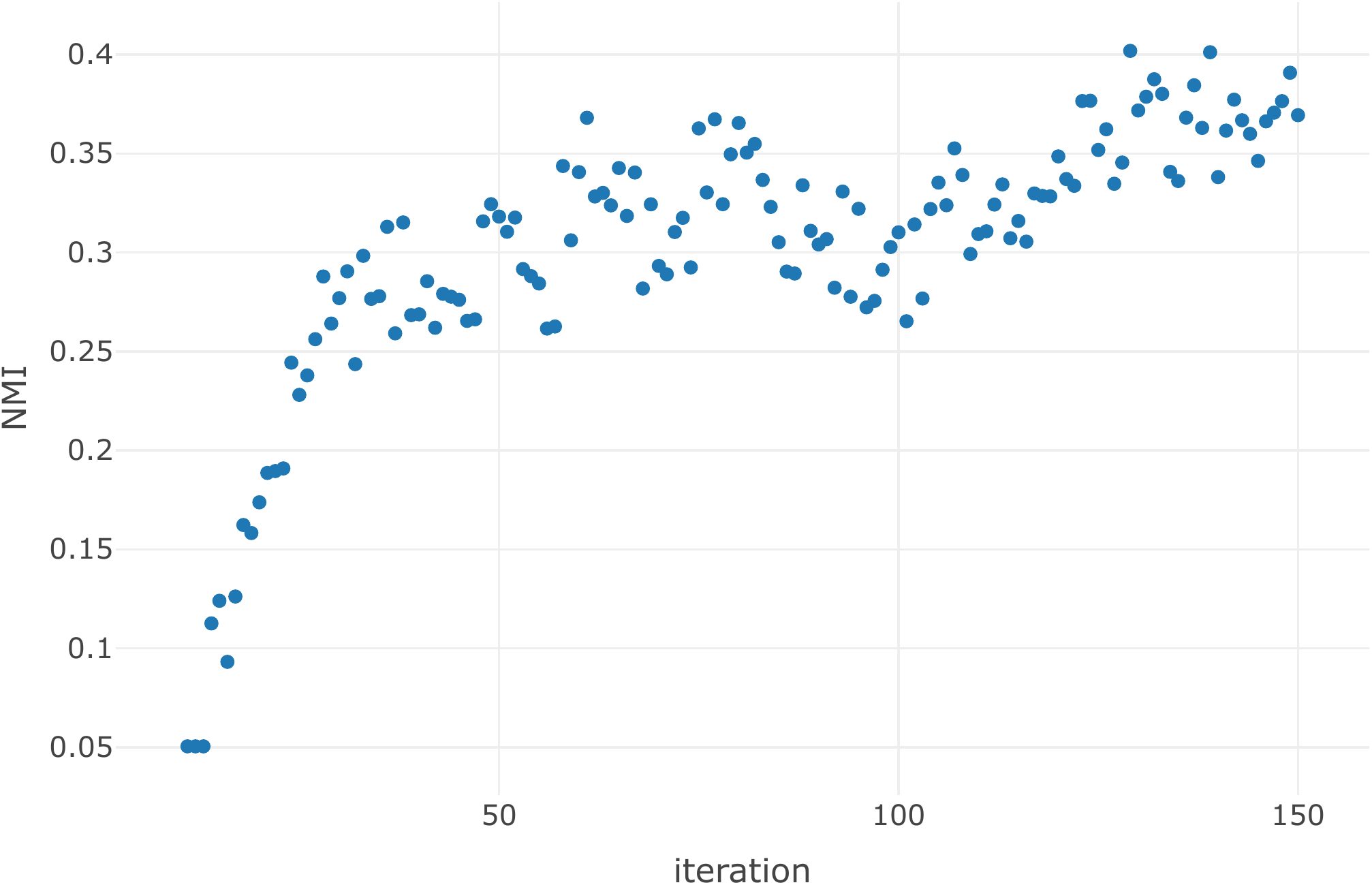}
	\includegraphics[width=.49\textwidth]{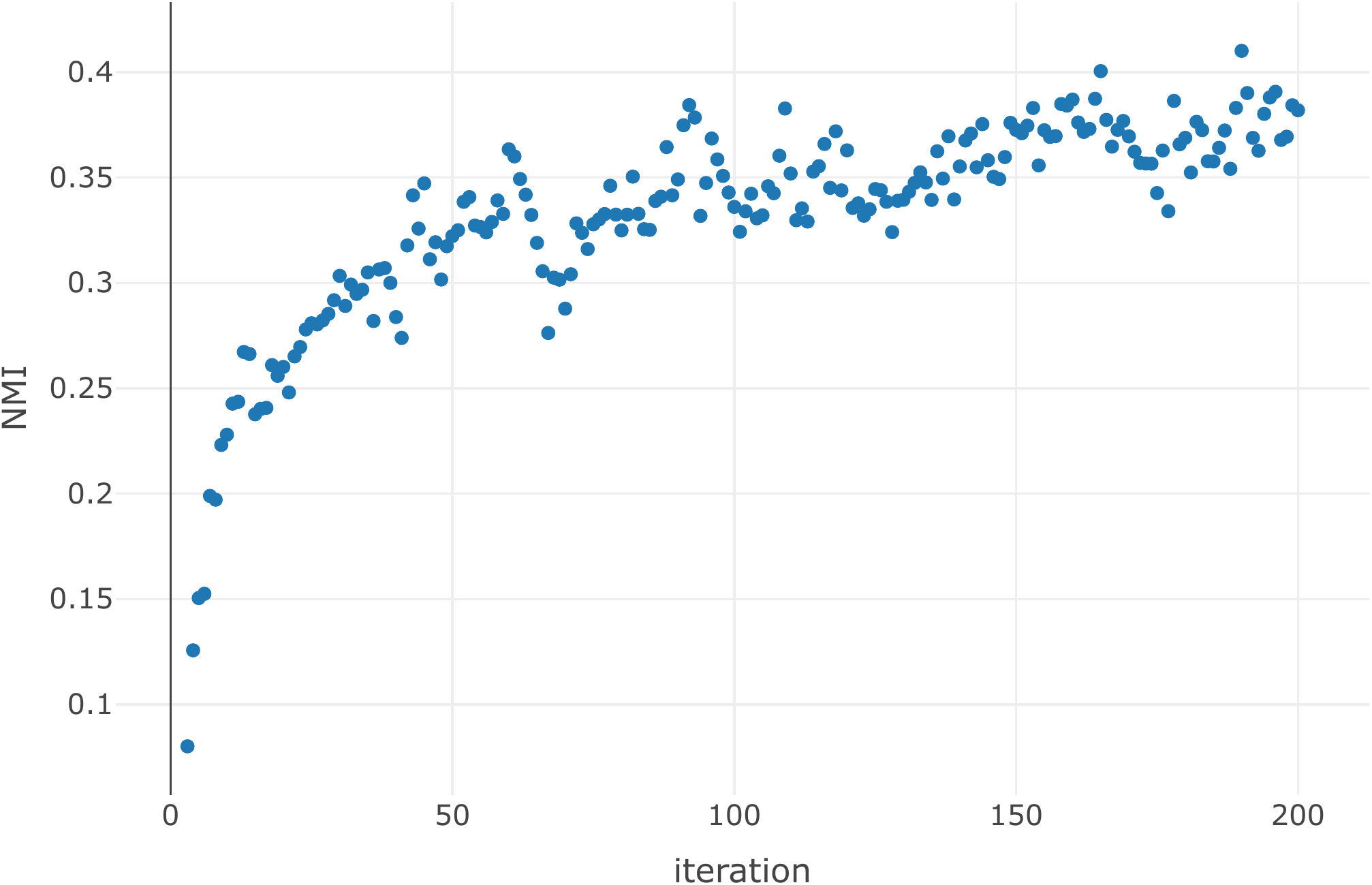}
	\caption{Results for a real data experiment. The NMI (relative to true labels) versus iteration for the real-world paper-title network. (left) performance on a random subset of size $J=100$ of the data (right) on the whole dataset $J=894$.}
	\label{fig:real:world}
\end{figure}

\subsection{Experiments}\label{sec:experiments}

We now present some numerical experiments to illustrate the performance of the slice sampler.  Since HDP mixtures are very popular and their performance well-known, we will mostly focus on studying the mixing time of the sampler. 
 Figure~\ref{fig:mixing:1} illustrates the mixing behavior for the multinomial HDP mixture discussed in Section~\ref{sec:examples}. We have also experimented with the Gaussian mixtures but we omit them here due to similarity. The code for these experiments is available on GitHub, repository \href{https://github.com/aaamini/hdpslicer}{aaamini/hdpslicer}.

In each case we simulated from  HDP with concentration parameters $\gamma_0 = 3$ and $\alpha_0=1$ and have run the slice sampler on a single sample. %
The multinomial parameter $W$ is varied and we set $\alpha_w = 1/W$. In each case, we have $n_j = n$ for all $j$ and three values $n=30,100,300$ are considered. 
For simplicity, we have set the  number of restaurants to $J=W=10,20,50,200$.  
Figure~\ref{fig:mixing:1} illustrates single typical runs of the algorithm without burn-in or thinning; there is also no averaging over multiple runs and the labels are all initialized to 1 as in Algorithm~\ref{alg:hdp:slice:sampler}. 

We have calculated the normalized mutual information (NMI) between estimated $(z_{ji})$ and true labels $(z_{ji}^*)$, aggregated over all $(i,j)$. NMI measures the matching between two clusterings, its value being in $[0,1]$ with a value of 1 corresponding to a perfect match. Figure~\ref{fig:mixing:1} shows the quality of recovered labels relative to the true data-generating labels, over the iterations of the sampler. The plots clearly indicate a fast mixing time, somewhere between 10 to 20 iterations. We note the decrease in the variance of the posterior as $n$ increases which is expected.

We have also applied the algorithm to a real world example where the documents are papers and bag-of-word  are made from the words in their titles. A vocabulary of a total of $W=189$ was used after running standard text mining procedures for removing the stopwords, stemming, and so on. The information on a total of $J=894$ papers was collected from the DBLP website. The papers were published in 2017 in three  CS topics: machine learning, multimedia and security. We treated DBLP subject classification as the true cluster of each paper. The HDP mixture is run on the dataset which recovers a clustering for every word in each document. We then assign an estimated cluster to each paper by majority voting (among the estimated clusters for their words) and compare with the true labels.  Figure~\ref{fig:real:world} illustrates the resulting NMIs versus iteration. Both a random subset of the papers (with $J=100$) and the whole set is considered. Again, we observe that the algorithm is mixing very fast, and by about 100 iterations we already have pretty good quality labels (NMI $\in [0.35,0.4]$).

\printbibliography

\appendix
\section{Proof of Lemma~\ref{lem:basic:dp:slice}}\label{app:useful:lemma}
	Since $[F(\xb)]_j$ only depends on $x_1,\dots,x_{j-1}$, we have
	\begin{align*}
	p(x_j \mid \xb_{-j}) &\propto  b(x_j; \alpha,\beta) \prod_{i:\; z_i \,\ge\, j} [F(\xb)]_{z_i}\\
	&= b(x_j; \alpha,\beta) \prod_{i:\; z_i \,=\, j} [F(\xb)]_{j} \prod_{i:\; z_i \,>\, j } [F(\xb)]_{z_i} \\
	&\propto b(x_j; \alpha,\beta) \prod_{i:\; z_i \,=\, j} x_{j} \prod_{i:\; z_i \,>\, j } (1-x_j) \\
	&= b(x_j; \alpha,\beta) \; x_j^{\n_j(\zb)} (1-x_j)^{\n_{>j}(\zb)} 
	\end{align*}
	which gives the desired result.

\subsection{Remarks}
Due to the sick-breaking interpretation of $F(\xb)$, it is not hard to see that
\[
\sum_{\ell=1}^j [F(\xb)]_\ell + [P(\xb)]_j = 1,\quad \forall j \in \nats.
\]
where $P: [0,1]^\nats \to [0,1]^\nats$ is defined by $[P(\xb)]_j := \prod_{\ell \,\le\, j} (1-x_\ell)$. That is, $[F(\xb)]_j = x_j [P(\xb)]_{j-1}$ hence $[P(\xb)]_{j-1} < \tau$ implies $ [F(\xb)]_j < \tau$. In other words, 
\[\{j:\;[F(\xb)]_j \ge \tau \}\; \subseteq\; \{j:\,[P(\xb)]_{j-1} \ge \tau\}\]
and we can use the latter set to guarantee that we have enough atoms when truncating $F(\xb)$ at level~$\tau$. This is due to the fact the $j \mapsto [P(\xb)]_{j}$ is nonincreasing in $j$ as opposed to $j \mapsto [F(\xb)]_j$ which is not necessarily monotone. Note that $[P(\xb)]_{j-1}$ is easy to keep track of since it is the complement to the cumulative distribution associated with $F(\xb)$ up to index $j-1$. We take $[P(\xb)]_0 = 1$. 

\end{document}